\documentclass[11pt,a4paper]{article}
\usepackage{authblk}
\usepackage[]{ACL2020}
\usepackage{times}
\usepackage{latexsym}
\usepackage[T1]{fontenc}
\usepackage{graphicx}
\usepackage[utf8]{inputenc}
\usepackage{comment}

\usepackage{microtype}

\aclfinalcopy 

\title{LBMT team at VLSP2022-Abmusu: Hybrid method with text correlation and generative models for Vietnamese multi-document summarization}

\author[1]{Tan-Minh Nguyen*}
\author[1]{Thai-Binh Nguyen}
\author[1]{Hoang-Trung Nguyen}
\author[1]{Hai-Long Nguyen}
\author[1 2]{\\Tam Doan Thanh}
\author[3]{Ha-Thanh Nguyen}
\author[1]{Thi-Hai-Yen Vuong*}
\affil[ ]{\{20020081, 20020328, 20020083, long.nh, yenvth\}@vnu.edu.vn}
\affil[ ]{tamdt9@viettel.com.vn, nguyenhathanh@nii.ac.jp}

\affil[1]{VNU University of Engineering and Technology / Hanoi, Vietnam}
\affil[2]{Viettel Group / Hanoi, Vietnam}
\affil[3]{National Institute of Informatics / Tokyo, Japan}

\date{}

\begin{document}
\maketitle

\begin{abstract}
\label{sec:abstract}

Multi-document summarization is challenging because the summaries should not only describe the most important information from all documents but also provide a coherent interpretation of the documents. 
This paper proposes a method for multi-document summarization based on cluster similarity. 
In the extractive method we use hybrid model based on a modified version of the PageRank algorithm and a text correlation considerations mechanism. After generating summaries by selecting the most important sentences from each cluster, we apply BARTpho and ViT5 to construct the abstractive models. 
Both extractive and abstractive approaches were considered in this study. 
The proposed method achieves competitive results in VLSP 2022 competition~\cite{tran2022vlsp}.
\end{abstract}

\section{Introduction}
Summarization is an efficient way to process information in the context of big data, especially for non-expert users who want to get a quick overview of the most crucial ideas in a document \cite{radev2002introduction}.
This problem has been applied to many natural language processing tasks such as document retrieval, text classification, and question answering \cite{guo2020deep,mishra2021looking,nguyen2022transformer,vuong2022sm}.
With the increasing amount of online content, it is becoming increasingly difficult for users to find interesting and useful information. 

Multi-document summarization is the task of generating a summary from a collection of documents. 
More specifically, this paper considers the task of generating summaries from a collection of documents in Vietnamese. 
Multi-document summarization is a challenging task because the summaries should not only describe the most important information from all documents but also provide a coherent interpretation of the documents. 

In the past few years, several deep learning models have been proposed for the task of multi-document summarization. 
Extractive models are based on the idea of selecting sentences from the input documents and then combining the selected sentences to form the summary. 
Abstractive models are based on the idea of generating summary sentences from scratch. 
Abstractive models have shown better performance than extractive models but are more difficult to train.
The reason is that the abstractive models need to understand the content of the input documents and then generate summary sentences. 

Thanks to the development of pretrained language models, abstractive models have shown good performance in text summarization tasks. 
Pretrained language models are based on the idea of training a model on a large amount of unannotated data and then fine-tuning the model on the specific task and data set. 
Available pretrained language models in Vietnamese \cite{tran2021bartpho,phan2022vit5} allow us to apply these models to summarize Vietnamese text. 
 
This paper proposes a pipeline abstractive method for multi-document summarization that includes three main phases.
The first phase is single document extraction. In the second phase, we concatenate the candidate sentences from single documents. Then we perform a multi-document extraction to produce quickview summaries. In the third phase, we build multi-document abstractive models from quickview results.
The proposed method achieves competitive results in the VLSP 2022 competition.

\section{Related Work}
Multi-document summarization is an arduous problem due to some reasons including high redundancy as well as contradictory information inside the documents, sophisticated relation between documents, diverse input document types, etc. For tackling this problem, the model often is separated into two phases: the extractive phase for reducing redundancies and collecting important information,  abstractive phase for generating natural-language summarise.  In \cite{yasunaga2017graph}, the authors built a sentence-based graph where each node is a sentence and each edge is a relation between two sentences. Then they used a GRU model for extracting vector representation for each sentence and fed the graph and the representation vector to a GCN for aggregating the final representation vector. After the sentence embedding stage, a second-level GRU is employed for extracting the entire cluster's vector representation which will be employed with a sentence-level representation vector for estimating the salience of each sentence. For paying attention to the relation between words, sentences, and documents, the work in \cite{wang2020heterogeneous} has proposed HeterDoc-SUM Graph which represents each of these elements (word, sentence, and document) by a node. The sentences node and documents node are connected by the containing words node, which means word nodes have a role as the bridge between sentence and document node and enhance the relations that exist in the data including sentence-sentence, sentence-document, and document-document relation. The built graph is fed into a graph attention network. 

Due to the success of transformer-based models in recent years, this class of model that utilizes the attention-based mechanism has been employed for getting more abstract information about the context. The authors of \cite{jin2020multi} proposed a multi-granularity interaction network MGSUM based on the Transformer model for capturing the semantic relationships. Three levels of granularity include Word, Sentence, and Document. The sentence level is used for extractive summarization and the word level is utilized for abstractive summarising tasks. Additionally, the model performs the semantic information integrating and updating by a fusion-gate, concentrating on the important knowledge by sparse-attention mechanism. The Transformer models pre-trained on a large dataset also perform impressively on summarization tasks. The knowledge learned from large corpora can fill the gap of lacking training data for multi-document summarization. PEGASUS model introduced in \cite{zhang2020pegasus} utilized Transformer-based encoder-decoder architecture. The gap sentence generation mechanism has been used and shows that masking the sentence based on its importance can enhance the summarization ability of the model. The work in \cite{tu-etal-2022-uper} improved the extractive phase by utilizing the pre-trained phase's knowledge of the language model to calculate the word and sentence perplexity. This unsupervised method can be used for assessing the salience of the document, which can increase the effectiveness of the abstractive phase.

\section{Method}
Given the set of topic-related documents $D = \{d_i | i \in [1,N]\}$ where $N$ is the number of documents. Each document $d_i$ has $M_i$ sentences $\{s_{j,i} | j \in [1, M_i]\}$, where $s_{j,i}$ refers to the $j^{th}$ sentence in the $i^{th}$ document. The Absumu task aims to generate a concise and abstractive summary for the cluster of documents $D$. We tackle the problem following two approaches: extractive and abstractive summarization. The extractive summarization goal is to choose important sentences and phases from the original documents and combine them into a shorter form. The abstractive summarization aims to rewrite summaries based on the original documents. The generated summaries may contain new sentences and phases.

\subsection{Extractive approach}
We use Sentence BERT \cite{reimers2019sentence} with a pre-trained model PhoBERT \cite{nguyen2020phobert} following source code\footnote{https://github.com/DatCanCode/sentence-transformers} to present semantic vector of a sentence. Then we perform two methods to extract summary: similarity and TextRank.

\textbf{Text correlation} A document includes a title, anchor text, and news content. The authors write anchor text to summarize the content of a document. Thus, the correlation between anchor text and body text contributes to the summary. Our approach is to measure the Cosine similarity between every sentence in a document with its anchor text. Then we combine anchor text with top $M$ candidate sentences based on the Cosine similarity score to generate the summary of a document. Doing the same with other documents, we receive an extractive summary of a cluster. 

\textbf{TextRank} is a graph-based ranking model for text processing introduced in \cite{mihalcea2004textrank}. In this competition, we employ TextRank for sentence extraction. Particularly, for each cluster, we build a graph with a vertex as a sentence and an edge weight is the Cosine similarity between two vertexes it connects. Then, we apply an algorithm derived from PageRank \cite{brin1998anatomy} to calculate the score of a vertex. Sentences are sorted in reversed order of their scores, and top $N$ candidate sentences are selected.

\subsection{Abstractive approach}
Since both BARTpho \cite{tran2021bartpho} and ViT5 \cite{phan2022vit5} input limitation is 1024 tokens, these models can not handle the raw organizer's dataset. Thus, we first perform extractive summarization following the above methods to extract summaries. Then we create a new dataset consisting of (extractive summary, gold label) pairs to fine-tune BARTPho and ViT5 models.

\textbf{BARTpho} is the first large-scale monolingual sequence-to-sequence models pre-trained for Vietnamese. BARTpho has two versions: BARTpho-syllable and BARTpho-word. We use the BARTpho-word-base model with 150M params and word-level input text. The model is trained with (extractive summary, gold label) pairs dataset. Before inputting data into the model, we employ RDR-Segmenter from the VnCoreNLP library \cite{vu2018vncorenlp} to perform word segmentation.

\textbf{ViT5} is a pre-trained Transformer-based encoder-decoder model for the Vietnamese language. ViT5 has two versions: ViT5-large and ViT5-base. The model achieves state-of-the-art results on summarization in both Wikilingual and Vietnews corpus. We train the ViT5-base model with 225M params with (extractive summary, gold label) pairs dataset.

\section{Experiments}
\subsection{Metrics}
The official measures of the competition are ROUGE-2 scores and ROUGE-2 F1. The ROUGE-2 F1 score is calculated with formulas:

\begin{equation}
\resizebox{0.45 \textwidth}{!}{$ROUGE-2 \; P = \frac{|Matched \; N-grams|}{|Predict \; summary \; N-grams|}$}
\end{equation}
\begin{equation}
\resizebox{0.45 \textwidth}{!}{$ROUGE-2 \; R = \frac{|Matched \; N-grams|}{Reference \; summary \; N-grams|}$}
\end{equation}
\begin{equation}
\resizebox{0.45 \textwidth}{!}{$ROUGE-2 \; F1 = \frac{2 \times ROUGE-2 \; P \times ROUGE-2 \; R}{ROUGE-2 \; P + ROUGE-2 \; R}$}
\end{equation}

\subsection{Data analysis}
According to our statistics with the given training dataset, the average number of documents in a cluster is 3.105. The average and the maximum number of words in a document respectively are 715.23 and 3474. More particularly, Figure 1 below illustrates the distribution of the document's length in the training datasets.

\begin{figure}[ht]
    \centering
    \includegraphics[width=0.45\textwidth]{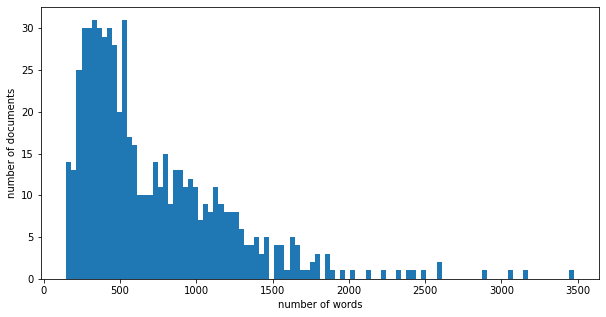}
    \caption{Document's length distribution}
    \label{fig:test document length distribution}
\end{figure}

Although the document's length varies from 150 to nearly 3500 words, the distribution mainly focuses on 250 to 750 words.

\begin{figure}[ht]
    \centering
    \includegraphics[width=0.45\textwidth]{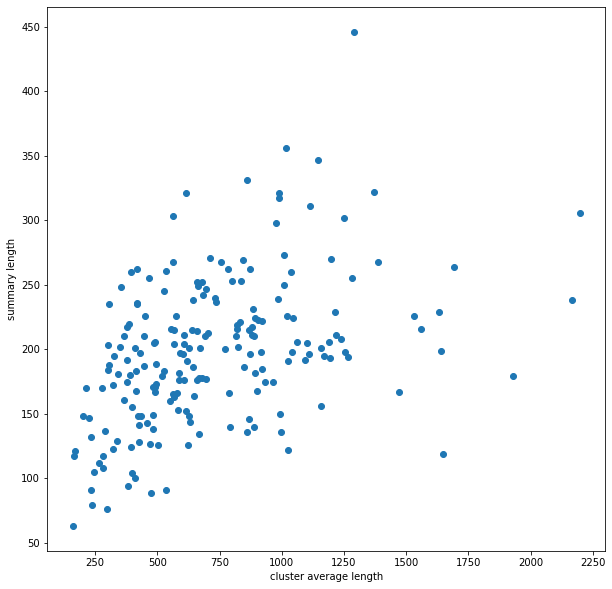}
    \caption{Cluster and summary lengths relation}
    \label{fig:train cluster and summary relation}
\end{figure}

\begin{figure}[ht]
    \centering
    \includegraphics[width=0.45\textwidth]{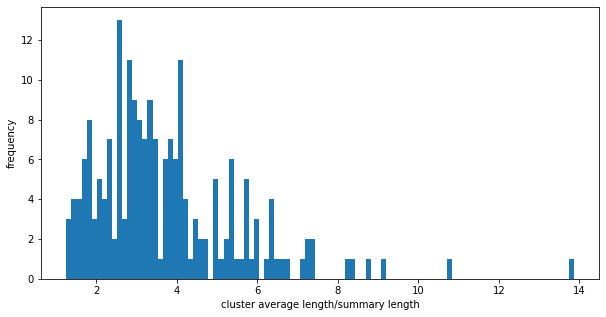}
    \caption{Ratio between cluster and summary lengths}
    \label{fig:train cluster and summary length ratio}
\end{figure}

We plot Figures \ref{fig:train cluster and summary relation} and \ref{fig:train cluster and summary length ratio} to estimate the summary's length based on cluster properties. The cluster average length is computed by the sum of all document's lengths divided by the number of documents. Based on Figure \ref{fig:train cluster and summary relation}, we argue that the cluster and summary's length have a linear relation. In Figure \ref{fig:train cluster and summary length ratio}, the ratio between a cluster and its summary's length is mainly 2.0 to 4.0.  

\subsection{Hyper-parameters setup}
All the experiments are performed on 16GB RAM and the support of GPU from Google Colab\footnote{https:colab.research.google.com}. 

\textbf{Text correlation} After measuring the Cosine similarity between every sentence and its anchor text in a document, we receive the candidate sentence list. For each document in a cluster, we take the top N sentences and combine them with their anchor text to generate a document's summary. Then we concatenate all the document summaries to form a cluster extractive summary. After multiple experiments, the top 3 sentence summary achieve the best ROUGE-2 F1 score.

\textbf{Multi extraction} We employ TextRank to build multi-document extractor. After multiple experiments with different settings, top $N$ candidate sentences with $N = 5$ achieves the best ROUGE-2 F1 score.

\textbf{BARTpho} We fine-tune the BARTpho model with (extractive summary, gold label) pairs dataset in 30 epochs. We set the minimum and maximum output lengths repspectively are 0.7 and 1 of their inputs. Since BARTpho is a generative model, it takes 1-2 minutes to generate a summary. We spent 6-7 hours producing 300 summaries of the testing dataset.

\textbf{ViT5} Because of the GPU limitation, we can only fine-tune the ViT5 model with (extractive summary, gold label) pairs dataset in 5 epochs. The minimum and maximum target lengths are respectively 0.7 and 1 of their input length. It takes 2-3 minutes for an iteration and approximately 2 hours to generate 100 abstractive summaries of the validation dataset.

\subsection{Results and Dicussion}

Table \ref{table: validation dataset results} shows our results on the validation dataset. We can see that the similarity method achieves the highest ROUGE-2 score. The ViT5 model's result is not good, so we only use the BARTpho model for the public and private testing datasets.

\begin{table}[ht]
\centering
\caption{Evaluation scores on validation dataset}

\begin{tabular}{|p{1.8cm}|p{1cm}|p{1cm}|p{1cm}|}
\hline
                  & R2-F1  & R2-P    & R2-R   \\ \hline
Multi extraction        & \textbf{0.4386} & \textbf{0.3498}  & \textbf{0.6560} \\ \hline
Text correlation          & 0.4249 & 0.4118  & 0.4640 \\ \hline
BARTpho-word-base & 0.4186 & 0.4452  & 0.4148 \\ \hline
ViT5-base         & 0.2734 & 0.3342 & 0.2308 \\ \hline
\end{tabular}
\label{table: validation dataset results}
\end{table}

\begin{table}[ht]
\centering
\caption{Evaluation scores on public test dataset}

\begin{tabular}{|p{1.8cm}|p{1cm}|p{1cm}|p{1cm}|}
\hline
                  & R2-F1  & R2-P    & R2-R   \\ \hline
Multi extraction        & \textbf{0.3149} & \textbf{0.2566}  & \textbf{0.4577} \\ \hline
BARTpho-word-base & 0.2607 & 0.2612  & 0.2911 \\
 \hline
Text correlation          & 0.2536 & 0.2094  & 0.3465 \\ \hline

\end{tabular}
\label{table: public test dataset}
\end{table}

Our team achieves second place in the public test leaderboard, less than the top 1 team 0.0001 score. Table \ref{table: public test leaderboard} shows the detailed results of the top 5 teams in the competition.

\begin{table}[ht]
\centering
\caption{Public test leaderboard}
\begin{tabular}{|p{2.2cm}|p{1cm}|p{1cm}|p{1cm}|}
\hline
User                & R2-F1               & R2-P                & R2-R                \\ \hline
thecoach\_team      & 0.3150        & 0.2492 & 0.4652       \\ \hline
\textbf{minhnt2709} & \textbf{0.3149} & \textbf{0.2566} & \textbf{0.4577} \\ \hline
TheFinalYear        & 0.2931        & 0.2424          & 0.4137         \\ \hline
TungHT              & 0.2875        & 0.2978         & 0.2989        \\ \hline
ngtiendong          & 0.2841         & 0.2908         & 0.2949         \\ \hline
\end{tabular}
\label{table: public test leaderboard}
\end{table}

We also have some discussion about our proposed method. The first limitation is that the proposed method cannot handle large document collections. The reason is that the TextRank algorithm is very slow, and it is impossible to use it for extensive document collections. The second limitation is that the proposed method does not consider the relationship between sentences because our goal in the competition is to achieve the highest ROUGE-2 score. However, it is possible to consider the relationship between sentences and the relationship between documents by changing the way the summary is generated. It would be an interesting topic for further research.

\section{Conclusion}
\label{sec:conclusion}

In this paper, we propose a pipeline abstractive method for multi-document summarization that includes three main phases.
The method uses text correlation, a modified version of the PageRank algorithm, and a generative model. 
Both extractive and abstractive approaches were considered in this study. 
The proposed method achieves competitive results in the VLSP 2022 competition. 
In future work, we will consider the relationship between sentences and the relationship between documents to improve the quality of summaries.

\bibliographystyle{acl_natbib}
\bibliography{ref}

\end{document}